\documentclass[twocolumn,twoside]{IEEEtran}
\usepackage[cmex10]{amsmath}
\usepackage{amsfonts,graphicx}
\usepackage[tight,footnotesize]{subfigure}
\usepackage{xcolor}
\definecolor{dark-red}{rgb}{0.4,0.15,0.15}
\definecolor{dark-blue}{rgb}{0.15,0.15,0.4}
\definecolor{medium-blue}{rgb}{0,0,0.5}
\usepackage[]{hyperref}
\hypersetup{pdftitle={Color image denoising by chromatic edges based vector valued diffusion}, 
    colorlinks, linkcolor={dark-red},
    citecolor={dark-blue}, 
    urlcolor={medium-blue}
}
\newcommand{\abs}[1]{\left\vert#1\right\vert}
\newcommand{\norm}[1]{\left\Vert#1\right\Vert}
\begin{document}
\title{Color Image Denoising by Chromatic Edges based Vector Valued Diffusion}
\author{V. B. Surya Prasath$^*$, Juan C. Moreno and K. Palaniappan
\thanks{$^*$Corresponding author. V. B. S. Prasath and K. Palaniappan are with the Department of Computer Science, University of Missouri-Columbia, MO 65211, USA, E-mail: \{prasaths,pal\}@missouri.edu. J. C. Moreno with the Instituto de Telecomunica\c{c}\~{o}es, Department of Computer Science, University of Beira Interior, 6201--001, Covilh\~{a}, Portugal, E-mail: jmoreno@ubi.pt.}}

\markboth{Preprint submitted to IEEE Signal Processing Letters}{Color image denoising by chromatic edges based vector valued diffusion}
\maketitle
\begin{abstract}
In this letter we propose to denoise digital color images via an improved geometric diffusion scheme. By introducing edges detected from all three color channels into the diffusion the proposed scheme avoids color smearing artifacts.  Vector valued diffusion is used to control the smoothing and the geometry of color images are taken into consideration. Color edge strength function computed from different planes is introduced and it stops the diffusion spread across chromatic edges. Experimental results indicate that the scheme achieves good denoising with edge preservation when compared to other related schemes.
\end{abstract}
\ifCLASSOPTIONpeerreview
\begin{center} \bfseries EDICS Category: IMD-ANAL \end{center}
\fi

\IEEEpeerreviewmaketitle

\section{INTRODUCTION}\label{intro}

In digital image restoration the primary task is to remove noise without changing the multiscale nature of images. There exists various methods for image denoising and variational-partial differential equations (PDE) based schemes~\cite{AK06} are powerful methods in this regard. In this paper we concentrate on anisotropic PDE based color image denoising and study how we can improve them by introducing color edges into the diffusion process. Perona and
Malik~\cite{PM90} proposed a nonlinear diffusion function to control the smoothing across edges for monochromatic images. Due to the efficient denoising property of such class of PDEs, recently there are efforts~\cite{KM00,TS01,BF02,BC08} to introduce them into color image denoising as well. Tschumperl\'{e} and Deriche~\cite{TD05} studied such diffusion PDEs in one common framework and unified them under vector valued regularization.

\begin{figure}
\centering
    \includegraphics[width=2.5cm]{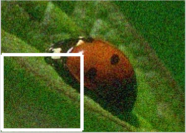}
    \includegraphics[width=2.5cm]{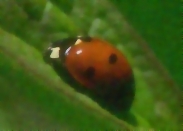}
    \includegraphics[width=2.5cm]{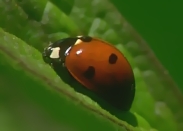}\\
    \subfigure[]{\includegraphics[width=2.5cm]{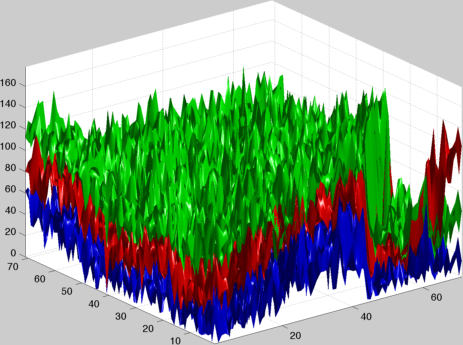}}
    \subfigure[]{\includegraphics[width=2.5cm]{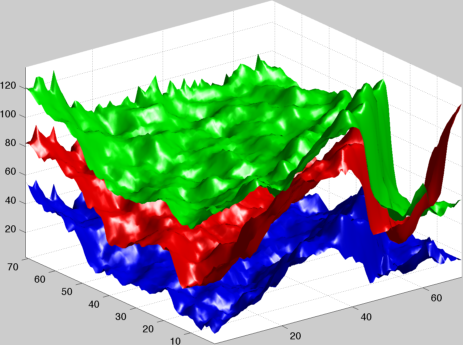}}
    \subfigure[]{\includegraphics[width=2.5cm]{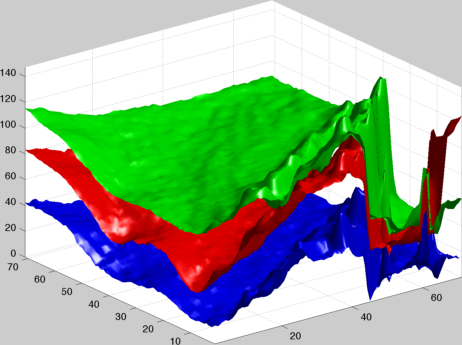}}
	\caption{Chromatic edges based diffusion PDE improves the denoising result. 
        (a) Input noisy image.
        (b) TD scheme~\cite{TD05}.
        (b) Proposed chromatic edges based vector valued diffusion scheme. Top row the images and bottom row shows the 3 color planes (RGB) of the cropped region indicated in top image in (a).}\label{I:bug}
        \vspace{-0.5cm}
\end{figure}

The main drawback of the current schemes is that they can introduce color smearing across multi-edges. Therefore low to medium contrast color edges are not preserved and the resultant image lacks clear color boundaries that are present in the original scene, see Figure~\ref{I:bug}. To avoid such artifacts we make use of chromatic edges found in different channels into the vector valued regularization of~\cite{TD05}. It is known that~\cite{CC02} the geometry of a color image is contained entirely in the base intensity image.  A weighted coupling is introduced to align and constraint the gradient responses of different color channels and the weights are computed from the input image. Thus, we make of use of not only the geometry but also the color information explicitly into the diffusion paradigm.

We introduce a color edge strength function via the coupling term into the vector valued regularization scheme. This function can be added to any of the other schemes mentioned as well and hence the proposed term can be utilized depending upon requirements. In this letter we consider denoising color images only, so a vector valued regularization based diffusion scheme is used. Apart from comparing our scheme with traditional diffusion based schemes~\cite{PM90,KM00,TS01,BC08,PS10c} we also compare our scheme with other coupling term based formulations~\cite{BF02,BK03} proposed in the past.

The paper is organized as follows. Section~\ref{prop} introduces the proposed scheme based on the vector valued diffusion model. Denoising results are presented in Section~\ref{deno}, and Section~\ref{concl} concludes the paper.
\section{PROPOSED SCHEME}\label{prop}

\subsection{Vector valued regularization}\label{vect}

Let $\mathbf{U}:\Omega\subset\mathbb{R}^2\to\mathbb{R}^3$ with $\mathbf{U}=(U_1,U_2,U_3)$ represent a color image in red, green, and blue (RGB) space. Given a noisy color image $\mathbf{U}_0$, the diffusion PDE of the following form,
\begin{eqnarray}\label{E:vecdiffn}
\frac{\partial U_i}{\partial t} = div\left(\mathbf{D}\cdot\nabla U_i(\mathbf{x},t)\right),~~\mathbf{x}\in\Omega \quad (i=1,2,3),
\end{eqnarray}
which is iteratively applied starting with the initial value $\mathbf{U}(x,0) = \mathbf{U}_0(x)$. The diffusion matrix  $\mathbf{D} $ is important in noise removal and is defined in terms of image gradients, see~\cite{AK06} for a review. Tschumperl\'{e} and Deriche (TD) in their work on vector valued image regularization~\cite{TD05} unified such diffusion PDEs in one common trace based framework,
\begin{eqnarray}\label{E:td}
\frac{\partial U_i}{\partial t} = trace\left(\mathbf{T}\mathbf{H}_i(\mathbf{x},t)\right),\quad (i=1,2,3).
\end{eqnarray}
where $\mathbf{H}_i$ is the Hessian matrix of the vector component $U_i$ and $\mathbf{T}$ is the tensor field defined point wise which we describe next. Consider the  smoothed multigradient, $K = G_{\sigma}\star\sum_{i=1}^3\nabla U^i(\nabla U^i)^T$ where $G_\sigma(\mathbf{x}) = (2\pi\sigma^2)^{-1}\exp{(-\abs{\mathbf{x}}/2\sigma^2)}$ is the Gaussian kernel, $\star$ denotes the convolution operation and superscript $T$ the transpose. Let the eigenvalues of $K$ be $\lambda_+$, $\lambda_{-}$, and their corresponding eigenvectors be $\theta_+$, $\theta_-$. If we let $\mathcal{N} = \sqrt{\lambda_+ + \lambda_-}$ which acts as an edge indicator in vector images, see for example Figure~\ref{I:chro}(a). 
\begin{eqnarray}\label{E:tensor}
\mathbf{T} = f_{-}(\mathcal{N})\,{\theta_-\theta^{T}_-} + f_+(\mathcal{N})\,{\theta_+\theta^{T}_+}
\end{eqnarray}
where $f_{-}(\mathcal{N}) = (1+(\mathcal{N})^2)^{-1/2}$ and $f_{+}(\mathcal{N}) = (1+(\mathcal{N})^2)^{-1}$.  Due to the use of multigradient $G$ and its eigenvectors, the overall geometry of the input vector image $\mathbf{U}$ is captured. Also, the channel coupling is implicitly integrated into the scheme by the eigenvalues in $\mathcal{N}$. Still smearing can occur near color edges which are not captured completely by the eigen decomposition of the multi-gradient $G$, see Figure~\ref{I:chro} bottom row. This can be attributed to the strong \textit{chromatic edges} which are specific to one channel and may or may not give a strong gradient response in other channels.

\subsection{Coupling based vectorial diffusion}\label{modi}
\begin{figure}
\centering
    \subfigure[]{\includegraphics[width=2.75cm]{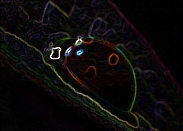}}
    \subfigure[]{\includegraphics[width=2.75cm]{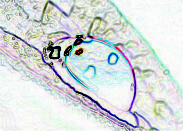}}
     \subfigure[]{\includegraphics[width=2.75cm]{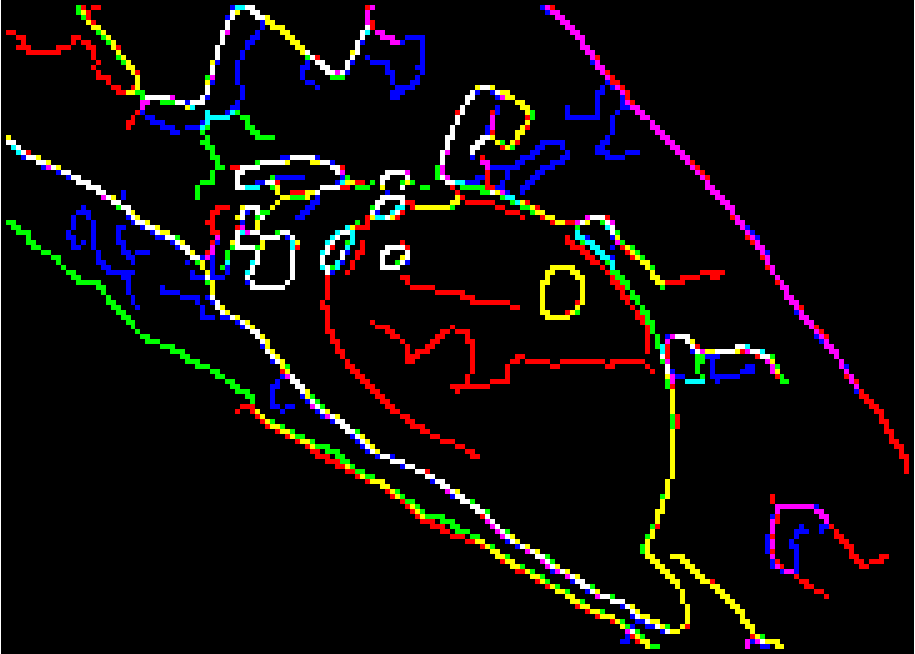}}
    \caption{Chromatic edges for the noisy image from Figure~\ref{I:bug}(a).
    (a) edge indicator function $\mathcal{N}$,
    (b) chromatic edges found by the coupling term~(\ref{E:chro}),
    (c) Canny edge detector based result for three color channels. The images are rescaled to $[0,1]$ range for visualization.}\label{I:chro}
     \vspace{-0.4cm}
\end{figure}
Following~\cite{Prasath11}, to control the color smearing we incorporate chromatic edges into the regularized PDE~\eqref{E:td} via weighted Laplacian computed from all three color channels. For this purpose we consider a coupling term of the form,
\begin{eqnarray}\label{E:chro}
f_C (U_i)=\sum_{j=1}^3\left(\omega_i\Delta U_j-\omega_j\Delta U_i\right) \quad (i=1,2,3),
\end{eqnarray}
where $\Delta$ is the Laplacian and the $\omega_i$ are the weight corresponding to channel $i$ to be modeled next.
The above equation facilitates an explicit coupling between different color channels and chromatic edges specific from each of the channel is adjusted to align themselves. The weights $\omega_i$ are added to adjust the magnitude of chromatic edges so as to avoid data mixup artifacts when computing the vectorial diffusion. In this letter, we choose the weights using the scalar total variation (TV) PDE~\cite{RO92},
\[\frac{\partial \tilde U^i}{\partial t} = div\left(\frac{\nabla \tilde U_i}{\abs{\nabla \tilde U_i}}\right)\quad (i=1,2,3),\]
applied channel-wise starting with the initial value $\mathbf{\tilde U}(\mathbf{x},0) = \mathbf{U}_0(\mathbf{x})$, then the weights are set to,
\begin{eqnarray}\label{E:weight}
\omega_i = \left(\abs{G_\rho\star\norm{\nabla \tilde
U_i}}\right)/\left(\sum_{j=1}^{3}\abs{G_\rho\star\norm{\nabla \tilde U_j}}\right)\quad (i=1,2,3),
\end{eqnarray}
where $\norm{\nabla \xi} = \sqrt{\xi_x^2 + \xi_y^2}$. 
Thus, the weights adjusts chromatic edges in terms of magnitude of gradients and aligns different channel edges via Laplacian differences in Eqn.~\eqref{E:chro}. The pre-smoothing with Gaussian $G_\rho$ is added to remove small oscillations which can occur due to the application of TV as well as to avoid outliers in gradient computations. Figure~\ref{I:chro} shows the importance of the chromatic edges of the image Figure~\ref{I:bug}(a). Thus, the scheme we propose takes the following form
\begin{eqnarray}\label{E:prop}
\frac{\partial U_i}{\partial t} = trace\left(\mathbf{T}\mathbf{H}_i(\mathbf{x},t)\right) +  \sum_{i=1}^{3} f_{C}(U_i)
\end{eqnarray}
Note that the weighted coupling term~\eqref{E:chro} can be added to any of the vector diffusion PDEs based on Perona-Malik diffusion term~\cite{PS10c} or in general to any diffusion PDE which computes intra-channel smoothing.
The pre-smoothing parameter $\sigma$ can be chosen according to the amount of noise present in the input image. If $\sigma$ is chosen very big, we lose the small scale chromatic edges. Similarly, the number of iterations to run the TV PDE depends on the noise level, and can be fixed a priori which could work for a class of images. By incorporating the weighted Laplacian differences between different channels we align the high frequency content while denoising with vector diffusion PDE. Figure~\ref{I:chro}(b) shows the chromatic edges found by the total coupling term $\sum_if_C$ along with Canny edge detector\footnote{MATLAB command: \texttt{\href{http://www.mathworks.com/help/images/ref/edge.html}{edge}($U_i$,`Canny',thresh,$\sigma$)}.} based result in Figure~\ref{I:chro}(c) for comparison. Other choices for the weight are possible, we chose the smoothed gradients as it gives a road-map to adjust and match the magnitudes of edges.
\section{EXPERIMENTAL RESULTS}\label{deno}

All the images are normalized to the range $[0,1]$ and we implement~\eqref{E:prop} by standard finite difference scheme. The algorithm is visualized in MATLAB 7.8(R2009a) on a 64-bit windows laptop with 3Gb RAM, 2.20GHz. For the TV PDE based smoothed weights computation~\eqref{E:weight} split-Bregman fast iteration scheme~\cite{GS09} is utilized ($<1$ \textit{sec} for $50$ iterations for $3$ channels). The pre-smoothing parameter in~\eqref{E:weight} and in the multigradient $K$ is set to $\sigma =$$\rho= 2$ and the TV PDE is run for a fixed $50$ iterations in all the examples reported here.

\begin{figure}
\centering
    \subfigure[Noisy]{\includegraphics[width=2.75cm]{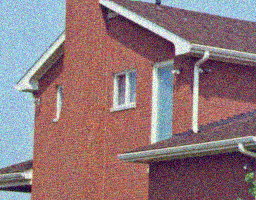}
    \includegraphics[width=2.75cm]{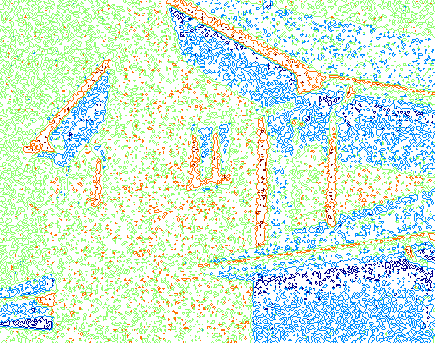}
    \includegraphics[width=2.75cm]{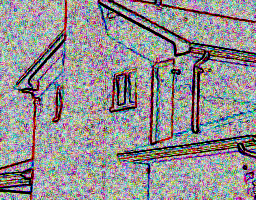}}
    \subfigure[TD scheme~\cite{TD05}]{\includegraphics[width=2.75cm]{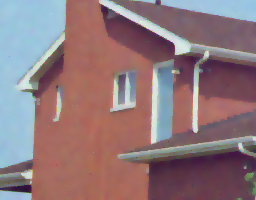}
    \includegraphics[width=2.75cm]{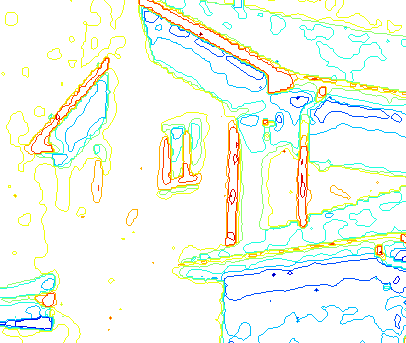}
    \includegraphics[width=2.75cm]{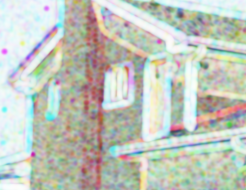}}
    \subfigure[Our scheme Eqn.~\eqref{E:prop}]{\includegraphics[width=2.75cm]{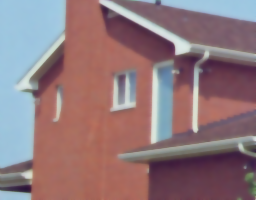}
    \includegraphics[width=2.75cm]{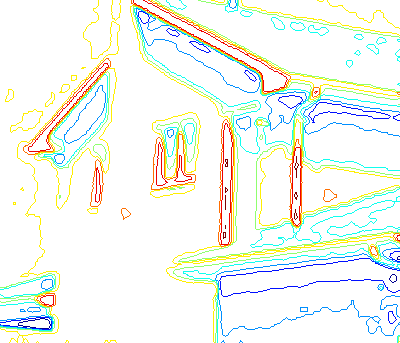}
    \includegraphics[width=2.75cm]{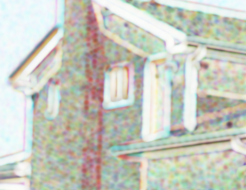}}
    \caption{Comparison with vectorial diffusion scheme.
    (a) Noisy $House$ image with its contour (middle) and combined edge map (right).
    (b) TD scheme~\cite{TD05}.
    (c) Our scheme.
    The proposed scheme retains the geometrical properties of the vectorial diffusion as can be seen from the contour (middle) and SSIM maps (right) in (b)\&(c). Better viewed online and zoomed in.}\label{I:bab}
    \vspace{-0.5cm}
\end{figure}

The computational time depends on the number of iterations of the main PDE~\eqref{E:prop} and for a $512\times 512$ RGB color image it takes less than a minute. The first example in Figure~\ref{I:bab} compares the denoising result of a close-up $House$ color image with the original vectorial diffusion TD~\cite{TD05} scheme. The noisy image given in Figure~\ref{I:bab}(a) is obtained by adding additive Gaussian noise of $\sigma = 20$ to the three color channels (R, G, B) separately. By comparing Figure~\ref{I:bab}(b) with Figure~\ref{I:bab}(c) we see that the proposed modified scheme preserves edges without color smearing near edges as well as avoids chromatic noise, see the contour maps in the middle.

Next we compare the performance of the proposed scheme~\eqref{E:prop} with the following schemes: anisotropic diffusion (PM, using finite differences)~\cite{PM90}, Beltrami flow (KMS, finite differences)~\cite{KM00}, generalized chromatic diffusion (BFT, finite differences)~\cite{BF02} which is an improvement over the chromatic diffusion scheme (TSC)~\cite{TS01}, variational regularization (BKS, gradient descent with finite differences)~\cite{BK03}, vector valued regularization of (TD\footnote{\href{https://tschumperle.users.greyc.fr/softwares.php}{https://tschumperle.users.greyc.fr/softwares.php}})~\cite{TD05}, vectorial TV (BC, dual minimization\footnote{\href{http://www.cs.cityu.edu.hk/~xbresson/codes.html\#fastcolor}{http://www.cs.cityu.edu.hk/~xbresson/codes.html\#fastcolor}})~\cite{BC08}, Perona and Malik with a coupling term~\cite{PS10c} (PS, finite differences). 

\begin{figure}
\centering
    \subfigure[Original]{\includegraphics[width=2.8cm]{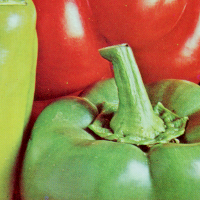}}
    \subfigure[Noisy]{\includegraphics[width=2.8cm]{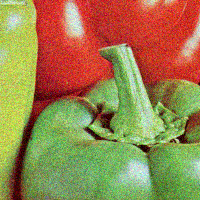}}
    \subfigure[Edges]{\includegraphics[width=2.8cm]{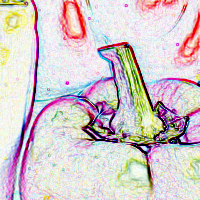}}
    \subfigure[PM~\cite{PM90}]{\includegraphics[width=2.8cm]{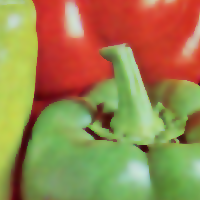}}
    \subfigure[KMS~\cite{KM00}]{\includegraphics[width=2.8cm]{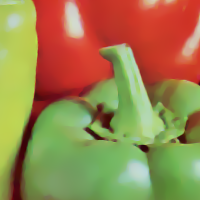}}
     \subfigure[TSC~\cite{TS01}]{\includegraphics[width=2.8cm]{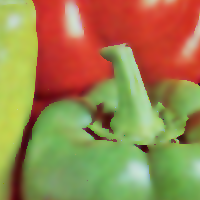}}
    \subfigure[BFT~\cite{BF02}]{\includegraphics[width=2.8cm]{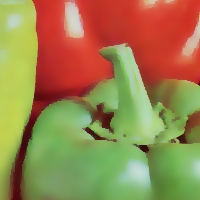}}
    \subfigure[BKS~\cite{BK03}]{\includegraphics[width=2.8cm]{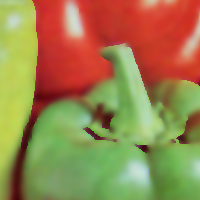}}
    \subfigure[TD~\cite{TD05}]{\includegraphics[width=2.8cm]{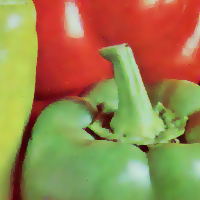}}
    \subfigure[BC~\cite{BC08}]{\includegraphics[width=2.8cm]{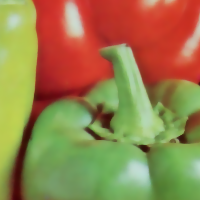}}
    \subfigure[PS~\cite{PS10c}]{\includegraphics[width=2.8cm]{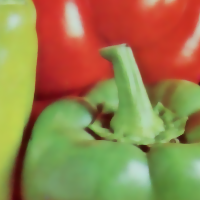}}
    \subfigure[Our scheme]{\includegraphics[width=2.8cm]{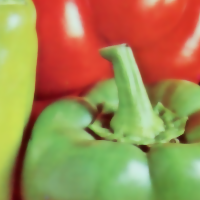}}
    \caption{Comparison results. Closeup of result for $Peppers$ color image for various schemes and chromatic edges computed with $\sigma=2$ used in our scheme is shown in (b).}\label{I:pepc}
    \vspace{-0.6cm}
\end{figure}

Figure~\ref{I:pepc} shows a comparison of the schemes output with our scheme result on a closeup (size $200\times 200$) of noisy $Peppers$ test image with $\sigma_n=20$. As can be seen from the images, by visual comparison, color smearing artifacts appear near the chromatic edges in PM, BFT and KMS scheme results due to wrong channel mixing and diffusion transfer. On the other hand BKS and BC scheme suffer from staircasing artifacts in flat regions.  The parameters involved in these schemes were chosen according to maximum peak signal-to-noise ratio (PSNR) value. PSNR for an estimated 3-dimensional color image $\mathbf{\hat U}$ is given by $\text{PSNR} (\mathbf{\hat U})  = 10\,
\log{(3/\text{MSE})}\, dB$ where $\text{MSE} = \sum_{x\in\Omega}(\mathbf{\hat U}(x)-\mathbf{U}(x))^2/\abs{\Omega}$ is the mean squared error with $\mathbf{U}$ the original noise-free image, $\abs{\Omega}$ is the dimension of the image domain $\Omega$.
To further assess the performance of different schemes we utilize the Mean Structural SIMilarity (MSSIM)~\cite{WB04} which is better suited than that of PSNR in terms finding the structure preservation~\footnote{\href{https://ece.uwaterloo.ca/~z70wang/research/ssim/index.html}{https://ece.uwaterloo.ca/~z70wang/research/ssim/index.html}}. Figure~\ref{I:bab}(b) \& (c) top right part shows the combined SSIM map of the restored results for the TD and our scheme.

Finally, Table~\ref{t:psnr} \& Table~\ref{t:mssim} compare the PSNR and MSSIM values achieved by various regularization and PDE based schemes on standard data set of natural color images of size $256\times 256$ and $512\times 512$, we show results for $17$ images and the remaining results, data-sets, denoised images are available at figshare\footnote{\href{http://dx.doi.org/10.6084/m9.figshare.658958}{http://dx.doi.org/10.6084/m9.figshare.658958}} and multispectral extension of the proposed scheme is straightforward. We conclude that the proposed chromatic edges based vectorial diffusion scheme performs better than other diffusion schemes and further improves color enhancement property of Eqn.~\eqref{E:td} with multi-edge preservation. Note that, for images with fine scale textures (see entries for $Baboon$, $Tree$ in Table~\ref{t:mssim}) our scheme has slightly lesser MSSIM values than TD scheme since capturing color textures are much harder than strong chromatic discontinuities. Extensions of the proposed denoising scheme for other videos~\cite{MS05}, image decomposition~\cite{AKb06} and demosaicking~\cite{Dubois05} can also be considered.
\begin{table*}
\caption{PSNR ({d}B) values for results of various schemes with noise level $\sigma_n=20$ for standard images of size $\dagger = 256\times 256$ (Noisy $\text{PSNR} = 22.11$) and $\ddagger = 512\times 512$ (Noisy $\text{PSNR} = 22.09$). Higher PSNR value indicate better denoising result. The best results for each image is indicated with boldface and the second best is underlined.}\label{t:psnr} \centering
    \begin{tabular}{llllllllll}
        \hline
Image &PM~\cite{PM90}& KMS~\cite{KM00}&TSC~\cite{TS01}& BFT~\cite{BF02} & BKS~\cite{BK03} &  TD~\cite{TD05} & BC~\cite{BC08}& PS~\cite{PS10c}&  Our\\
        \hline
            Baboon$\ddagger$  	& 21.53 &  21.75 & 21.21 & \underline{21.84} & 20.53 & \textbf{23.53} & 16.08 & 19.51 & 21.58\\
            Barbara$\ddagger$ 	& 26.13 & 26.60 & 25.44 & 25.95 & 24.94 & \textbf{27.27} & 20.36 & 24.47 & \underline{26.69}\\
            Boat$\ddagger$  	  	& 25.05 & 25.29 & 24.35 & 25.21 & 23.34 & \textbf{26.85} & 18.99 & 23.90 & \underline{25.81}\\
	  Car$\ddagger$         	& 26.23 & \underline{26.85} & 25.35 & 26.07 & 24.68 & \textbf{27.72} & 18.86 & 23.64 & 25.97\\
	  Couple$\dagger$    	& 28.41 & 28.98 & 27.68 & 28.10 & 26.97 & 29.69 & 27.72 & \underline{29.21} & \textbf{29.81}\\
            F-16$\dagger$   	 	& 26.62 & \underline{27.20} & 25.90 & 26.85 & 24.65 & \textbf{28.50} & 18.77 & 26.78 & 25.42\\
            Girl1$\dagger$        	& 28.50 & 28.76 & 27.78 & 27.96 & 27.24 & \underline{29.56} & 24.37 & 27.99 & \textbf{29.71}\\
            Girl2$\dagger$ 	 	& 31.18 & \underline{32.11} & 30.38 & 31.16 & 29.55 & \textbf{32.59} & 18.63 & 23.25 & 27.49\\
            Girl3$\dagger$ 		& 29.80 & 30.74 & 28.63 & 29.39 & 27.68 & \underline{31.04} & 19.72 & 24.49 & 28.20\\
            House$\dagger$		& 28.90 & \textbf{30.20} & 27.92 & 28.93 & 26.85 & \underline{30.18}& 17.92 & 24.81 & 29.76\\
            IPI$\dagger$			& 31.54 & \underline{32.63} & 30.34 & 31.30 & 29.13 & \textbf{32.72} & 23.72 & 27.97 & 21.27\\
            IPIC$\dagger$		& 29.93 & \underline{31.22} & 28.55 & 29.55 & 27.19 & 31.13 & 21.45 & 28.34 & 22.15\\
            Lena$\ddagger$ 		& 28.80 & \underline{29.44} & 28.00 & 28.44 & 27.39 & \textbf{29.84} & 17.83 & 22.52 & 26.85\\
            Peppers$\ddagger$	& 29.27 & \underline{29.80} & 28.45 & 28.91 & 27.81 & \textbf{30.07} & 21.94 & 29.09 & 28.89\\
            Splash$\ddagger$  	& 32.91 & \textbf{33.80} & 31.87 & 32.65 & 31.11 & \underline{33.52} & 18.85 & 24.90 & 31.62\\
            Tiffany$\ddagger$		& 29.87 & \underline{30.22} & 29.38 & 29.68 & 28.91 & \textbf{30.88} & 13.17 & 17.78 & 22.58\\
            Tree$\dagger$   		& 25.05 & \underline{25.92} & 24.20 & 25.23 & 23.10 & \textbf{26.64} & 18.99 & 23.67 & 25.73\\
        \hline
    \end{tabular}
\end{table*}
\begin{table*}
\caption{MSSIM values for results of various schemes with noise level $\sigma_n=20$ for standard images (size $\dagger = 256\times 256$ and $\ddagger = 512\times 512$). MSSIM value
closer to one indicates the higher quality of the denoised image. The best results for each image is indicated with boldface and the second best is underlined.}\label{t:mssim} \centering
    \begin{tabular}{llllllllll}
        \hline
Image &PM~\cite{PM90}& KMS~\cite{KM00}&TSC~\cite{TS01}& BFT~\cite{BF02} & BKS~\cite{BK03} &  TD~\cite{TD05} & BC~\cite{BC08}& PS~\cite{PS10c}&  Our\\
        \hline
            Baboon$\ddagger$	& 0.7023 & 0.7073 & 0.6676 & 0.6887 & 0.6047 & \textbf{0.8225} & 0.5259 & 0.6815 & \underline{0.7594}\\
            Barbara$\ddagger$	& 0.8358 & 0.8411 & 0.8125 & 0.8189 & 0.7964 & \underline{0.8702} & 0.7792 & 0.8502 & \textbf{0.8812}\\
            Boat$\ddagger$  		& 0.7385 & 0.7440 & 0.7137 & 0.7313 & 0.6786 & \underline{0.7971} & 0.6943 & 0.7675 & \textbf{0.8024}\\
	  Car$\ddagger$    		& 0.8379 & 0.8484 & 0.8094 & 0.8214 & 0.7863 & \underline{0.8818} & 0.7766 & 0.8543 & \textbf{0.8880}\\
	  Couple$\dagger$ 		& 0.7225 & 0.7375 & 0.7000 & 0.7016 & 0.6825 & 0.7586 & 0.7210 & \underline{0.7746} & \textbf{0.8044}\\
            F-16$\dagger$   		& 0.7992 & 0.8121 & 0.7797 & 0.7985 & 0.7469 & \underline{0.8427} & 0.7747 & 0.8331 & \textbf{0.8593}\\
            Girl1$\dagger$ 		& 0.7550 & 0.7534 & 0.7367 & 0.7286 & 0.7247 & 0.7813 & 0.7212 & \underline{0.7840} & \textbf{0.8142}\\
            Girl2$\dagger$ 		& 0.8833 & \underline{0.8997} & 0.8739 & 0.8842 & 0.8652 & 0.8961 & 0.8553 & 0.8908 & \textbf{0.9093}\\
            Girl3$\dagger$ 		& 0.8445 & 0.8509 & 0.8237 & 0.8289 & 0.8114 & \underline{0.8609} & 0.8032 & 0.8551 & \textbf{0.8797}\\
            House$\dagger$ 		& 0.7704 & 0.7908 & 0.7519 & 0.7704 & 0.7308 & \underline{0.7930} & 0.7355 & 0.7844 & \textbf{0.8093}\\
            IPI$\dagger$			& 0.9250 & 0.9391 & 0.9159 & 0.9295 & 0.9039 & 0.9354 & 0.9306 & \underline{0.9493} & \textbf{0.9550}\\
            IPIC$\dagger$		& 0.9102 & 0.9257 & 0.8939 & 0.9093 & 0.8725 & 0.9253 & 0.9023 & \underline{0.9382} & \textbf{0.9478}\\
            Lena$\ddagger$		& 0.8710 & 0.8728 & 0.8550 & 0.8550 & 0.8420 & \underline{0.8936} & 0.7948 & 0.8748 & \textbf{0.9103}\\
            Peppers$\ddagger$	& 0.8964 & 0.8928 & 0.8860 & 0.8832 & 0.8762 & 0.9086 & 0.8656 & \underline{0.9211} & \textbf{0.9410}\\
            Splash$\ddagger$  	& 0.9078 & 0.9089 & 0.9050 & 0.9070 & 0.9020 & 0.9115 & 0.8687 & \underline{0.9230} & \textbf{0.9460}\\
            Tiffany$\ddagger$		& 0.8773 & 0.8773 & 0.8679 & 0.8639 & 0.8607 & \underline{0.8933} & 0.7833 & 0.8720 & \textbf{0.9103}\\
            Tree$\dagger$   		& 0.7056 & 0.7306 & 0.6786 & 0.7099 & 0.6382 & \textbf{0.7630} & 0.6754 & 0.7294 & \underline{0.7578}\\
             \hline
    \end{tabular}
    \vspace{-0.4cm}
\end{table*}
\vspace{-0.5cm}
\section{CONCLUSION}\label{concl}

Using the vector valued regularization method we propose a new color image denoising scheme based on color edges. The scheme performs edge preserving restoration to get better control of the chromatic part in a color image. Vector valued diffusion is used to control the geometry of the multichannel image and weighted coupling with total variation is used to align the chromatic edges. Experimental results with other related color image denoising schemes indicate the proposed approach improves the denoising capability. Color image decomposition with different color spaces and color texture preservation will be our next concern in this research direction.
\vspace{-0.4cm}
{\tiny 

}
\end{document}